%% file: main.tex
\DocumentMetadata{
      pdfversion=2.0,pdfstandard=ua-2,
      testphase={phase-III,firstaid,math,title}
    }

\documentclass[sigconf,screen]{acmart-tagged}
\usepackage{enumitem}
\usepackage[utf8]{inputenc}

\AtBeginDocument{%
  }

\setcopyright{cc}
\setcctype{by}
\acmDOI{10.1145/3776734.3794600}
\acmYear{2026}
\copyrightyear{2026}
\acmISBN{979-8-4007-2321-6/2026/03}
\acmConference[HRI Companion '26]{Companion Proceedings of the 21st ACM/IEEE International Conference on Human-Robot Interaction}{March 16--19, 2026}{Edinburgh, Scotland, UK}
\acmBooktitle{Companion Proceedings of the 21st ACM/IEEE International Conference on Human-Robot Interaction (HRI Companion '26), March 16--19, 2026, Edinburgh, Scotland, UK}
\received{2025-12-08}
\received[accepted]{2026-01-12}

\begin{document}

\title{RoboTales: ROBOTic Anthropomorphic LEarning Systems}

\author{Andrew Chen}
\orcid{0009-0000-2928-5538}
\affiliation{%
  \institution{Case Western Reserve University}
  \city{Cleveland}
  \country{USA}
}
\email{ahc80@case.edu}

\author{Ju-Hung Chen}
\orcid{0009-0001-4756-1441}
\affiliation{%
  \institution{Case Western Reserve University}
  \city{Cleveland}
  \country{USA}
}
\email{jxc2350@case.edu}

\author{Phurinat Pinyomit}
\orcid{0009-0005-8873-3528}
\affiliation{%
  \institution{Case Western Reserve University}
  \city{Cleveland}
  \country{USA}
}
\email{php34@case.edu}

\author{Alexis E. Block}
\orcid{0000-0001-9841-0769}
\affiliation{%
  \institution{Case Western Reserve University}
  \city{Cleveland}
  \country{USA}
}
\email{alexis.block@case.edu}






\renewcommand{\shortauthors}{Chen et al.}

\begin{abstract}
\input{sections/00_abstract}
\end{abstract}


\sloppy




\maketitle

\input{sections/01_introduction}
\input{sections/02_relatedworks}

\input{sections/03_goals}
\input{sections/design_context}
\input{sections/system_overview}

\input{sections/user_study}


\input{sections/06_results}
\input{sections/05_empower_society}

\vspace{-0.4cm}
\begin{acks}
\vspace{-0.1cm}
Thanks to the participants, Yui Ishihara, and Dr. Ben Richardson.\looseness-1
\end{acks}
\vspace{-0.4cm}
\bibliographystyle{ACM-Reference-Format}
\bibliography{refs}

\end{document}

%% file: sections/00_abstract.tex
RoboTales is a low-cost robotic storytelling system that animates narratives using expressive sock puppetry. Implemented autonomously on a Baxter robot as a test case, RoboTales synchronizes narration, gestures, and mouth movements to perform character-driven stories. In a pilot study, puppet-based storytelling outperformed a gesture-only mode, producing higher HRIES ratings and improved story recall, suggesting that embodied puppetry enhances engagement and narrative comprehension. Designed to be modular and platform-agnostic, RoboTales can be adapted to other manipulators and offers a screen-free alternative to passive media, supporting future deployment in child-centered learning environments.\looseness-1

%% file: sections/01_introduction.tex
\vspace{-0.1cm}
\section{Introduction}
\vspace{-0.1cm}
\begin{figure}
\centering
  \includegraphics[width=\linewidth,alt={Side-by-side images of two robots. (Left, A) A red bimanual robot with articulated arms, each fitted with a sock-puppet end-effector, and a simple cartoon face displayed on its screen. (Right, B) A pink robotic arm manipulator holding a single sock puppet. The puppets have stitched faces and yarn hair; one wears a hat and has mostly monotone colors, while the other features more colorful appearance.} trim={0cm 0cm 0cm 0cm},clip]{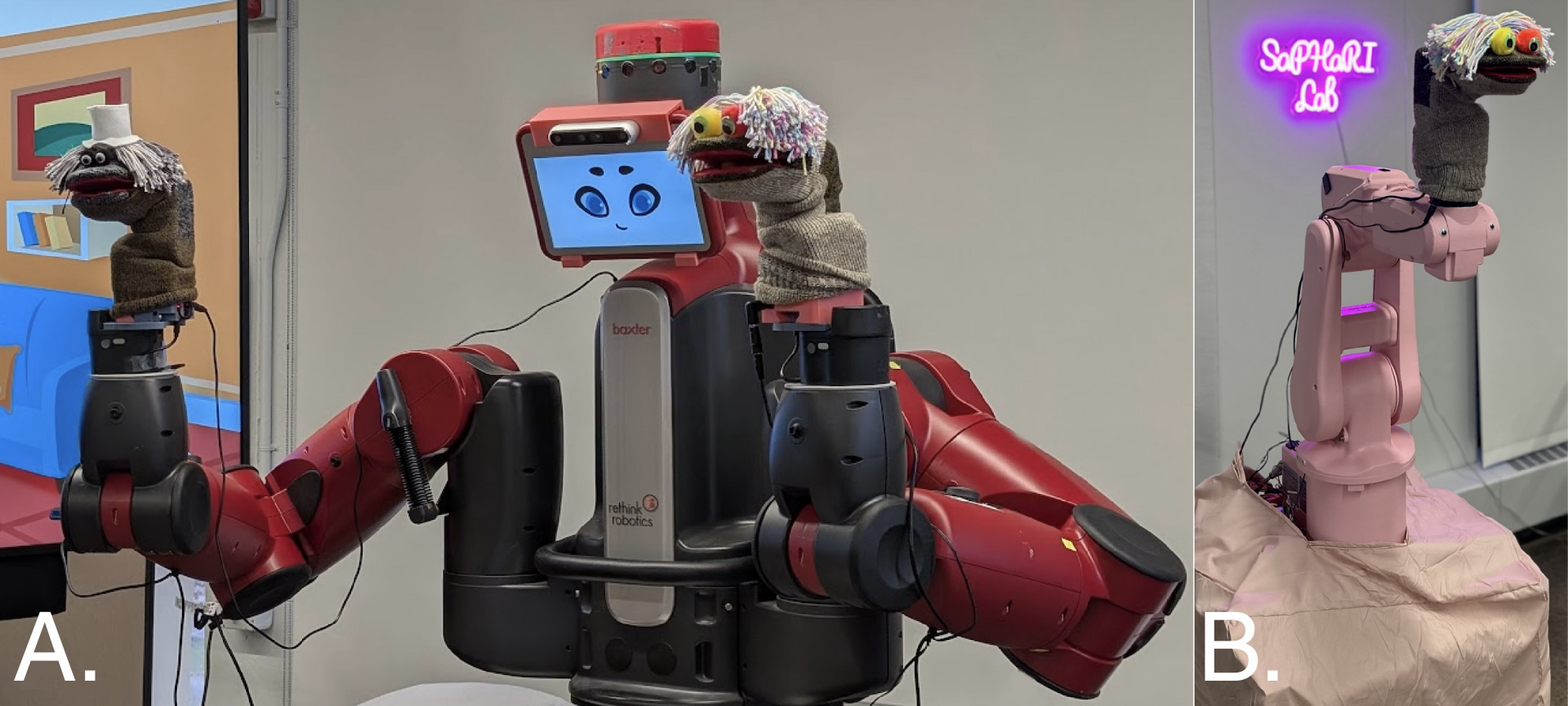}
  \vspace{-0.8cm}
  \caption{\normalfont{\textit{RoboTales performing a story using custom puppet end-effectors mounted on A.~a Baxter robot and B.~an ABB IRB 120 robot. RoboTales synchronizes puppet motion and narration to embody characters and deliver an expressive, autonomous storytelling experience.\looseness-1}}}
  \vspace{-0.55cm}
\label{fig:baxterwithpuppets}
\end{figure}
Puppetry has a long history as an educational and performative tool, using motion and character embodiment to simplify complex ideas~\cite{lambeth2019introduction}. In parallel, social assistive robots (SARs) are increasingly used to support learning and engagement across age groups~\cite{feil2005defining}. Both domains leverage embodiment and narrative, yet they have largely evolved separately.

Most SARs communicate only through speech, gaze, or gesture, modalities that struggle to convey emotional nuance, character identity, or narrative structure. In contrast, children respond strongly to narrative-driven characters~\cite{Aminimanesh_Ghazavi_Mehrabi_2019}, and puppetry has been shown to heighten attention, emotional engagement, and memory recall~\cite{mastrothanasis2025examining, luen2021puppetry}. However, existing robotic storytellers rely on screen-based avatars or rigid gesture libraries that limit engagement without supervision. \textit{RoboTales addresses this gap by enabling robots to autonomously control low-cost, physically embodied puppets for character-based storytelling in everyday educational environments.}

Because sock puppets require minimal actuation, the system can be mounted on a range of uni- or bimanual robots without specialized hardware, broadening accessibility and deployment. In this work, \textit{we contribute a reusable design pattern for embodied robotic storytelling} that integrates low-cost sock-puppet actuation, autonomous gesture generation, and synchronized narration to create expressive narratives without human oversight. 


%% file: sections/02_relatedworks.tex
\section{Related Work}
\vspace{-0.1cm}

Social robots are increasingly used to support learning and engagement, especially in childhood education~\cite{socialroboteducation, Lemaignan_Newbutt_Rice_Daly_2024}. Compared to screen-based systems, a robot's physical embodiment improves attention, memory, and social understanding~\cite{Yeung_Ma_Law_2025, Goldin-Meadow_2014}.\looseness-1


Puppets similarly leverage physical presence and recognizable character cues to support non-verbal communication, attention to social signals, and narrative comprehension~\cite{Karaolis_2023, Macari_Chen2021, Luen_2021}. Puppet-based storytelling has demonstrated behavioral, emotional, and attention benefits beyond non-embodied methods~\cite{Aminimanesh_Ghazavi_Mehrabi_2019, Macari_Chen2021}. Yet, most robot-led storytelling systems rely on human guidance or tablet-based displays~\cite{Yeung_Ma_Law_2025}, yielding engagement but inconsistent gains in comprehension and recall. \textit{This indicates a missing link between robotic embodiment and narrative expression.}

Robotic puppetry research has largely focused on complex marionettes~\cite{marionette1, johnson2007dynamic, Murphey_Argall}, which are difficult to deploy on general-purpose robots. In contrast, sock puppets require only simple mouth and neck articulations, are mechanically robust, and provide intuitive anthropomorphic cues~\cite{causo2015developing}. These properties make them well-suited for autonomous robotic storytelling in RoboTales, bridging accessible embodiment with narrative expressiveness.

%% file: sections/03_goals.tex
\section{Design Goals and Success Criteria}
\label{sec:goals}
\vspace{-0.1cm}
To ground RoboTales in practical storytelling needs and deployment constraints, we derived design goals from three ideas: 1)~limitations in prior robotic storytelling systems (limited embodiment and character expressivity), 2)~movement characteristics used in physical puppetry to convey identity, and 3)~realistic deployment constraints in community learning environments (low cost and ease of replication). These considerations motivate three design goals:
\vspace{-0.1cm}
\begin{enumerate}[nosep]
    \item[\textbf{G1:}] The system should synchronize narration with physical embodiment (mouth, head motion, and gestures) to convey character identity, emotion, and narrative meaning. 
    \item[\textbf{G2:}] The system should use affordable, maintainable components that non-experts can easily fabricate, operate, and repair.
    \item[\textbf{G3:}] The system should support multiple stories/characters and transfer across robot platforms through modular hardware and software pipelines. 
\end{enumerate}
To assess these goals, we defined success criteria for each design goal and evaluated them in a small pilot study:
\vspace{-0.1cm}
\begin{enumerate}[nosep]
    \item[\textbf{S1:}] Viewers should demonstrate engagement and narrative comprehension, evidenced by HRIES and story recall responses.
    \item[\textbf{S2:}] The system should run autonomously with off-the-shelf components and require little to no human intervention.
    \item[\textbf{S3:}] The puppet hardware and software pipeline should be adaptable with minimal modification, enabling flexibility across different robotic platforms and narratives.
\end{enumerate}

These goals and success criteria will ensure RoboTales can serve as a reusable design pattern for embodied robotic storytelling.

%% file: sections/design_context.tex
\section{Design Context}
\vspace{-0.1cm}
\textbf{\textit{Who is the human?}} RoboTales is designed for \textit{children and early learners}, and the educators who facilitate reading, narrative learning, and creative play. It can also engage adult audiences in informal learning environments, like libraries and community workshops.

\vspace{0.05cm}
\noindent \textbf{\textit{Key Characteristics and Robot Details.}} To evaluate RoboTales, we use a Baxter robot (Fig.~\ref{fig:baxterwithpuppets}), as a test case, chosen for its expressive 7-DoF arms and safe, collaborative design. The system adds two custom-built, \textit{two-DoF sock-puppet end-effectors}, enabling mouth articulation and simple head movements. These lightweight puppets provide clear anthropomorphic cues while remaining mechanically simple, inexpensive, and durable. Because the design requires only minimal actuation, the puppet end-effectors can be mounted on any uni- or bimanual robot, expanding accessibility.

\vspace{0.05cm}
\noindent \textbf{\textit{Activity.}} The robot performs \textit{interactive storytelling} using synchronized gestures, audio, and puppet mouth movements to enact short narratives. The human observes, listens, and responds to the story through laughter, curiosity, verbal comments, or emotional engagement. Educators can tailor the story themes to support literacy, emotional development, cultural traditions, or learning activities.

%% file: sections/system_overview.tex
\section{System Overview}
\vspace{-0.1cm}
RoboTales combines physical design, simple actuation, and coordinated motion to create expressive robotic puppetry by creating custom-designed, 3D-printed puppet end-effectors. 

We translated our design goals into concrete physical constraints that support puppet-like storytelling, while keeping the system accessible for real-world use. To satisfy G1 without unnecessary mechanical complexity, we implemented a minimal two DoF puppet end-effector: a)~a servo-driven jaw that renders narration as mouth articulation, and b)~a servo-driven neck that enables nodding to convey emphasis and character affect. To support G2, the mechanism uses affordable off-the-shelf servos and 3D-printed parts that are \textit{easy to fabricate, repair, and reuse} in educational settings. This modular design also supports G3 by allowing rapid swapping of puppet characters and transfer to other robot arms.

The end-effectors were fully CAD-modeled in SolidWorks and 3D-printed in PLA on a Bambu Lab X1C 3D printer, allowing rapid iteration and low-cost replication. 

\vspace{-0.2cm}
\subsection{Hardware}
\vspace{-0.1cm}
\begin{figure}[b]
    \centering
    \vspace{-0.6cm}
    \includegraphics[width=0.7\linewidth,alt={Four images arranged in a 2×2 grid. (Top left, A) A gray CAD rendering of a compact robotic end-effector with jointed components. (Top right, B) A pink 3D-printed robotic end-effector with visible Dynamixel servos and labeled PLA structure. (Bottom left, C) A sock puppet named Barneby with multicolored yarn hair, mismatched button eyes, and an open red mouth. (Bottom right, D) A sock puppet named Fitzwilliam with gray yarn hair, round button eyes, a small white hat, and an open red mouth.} ]{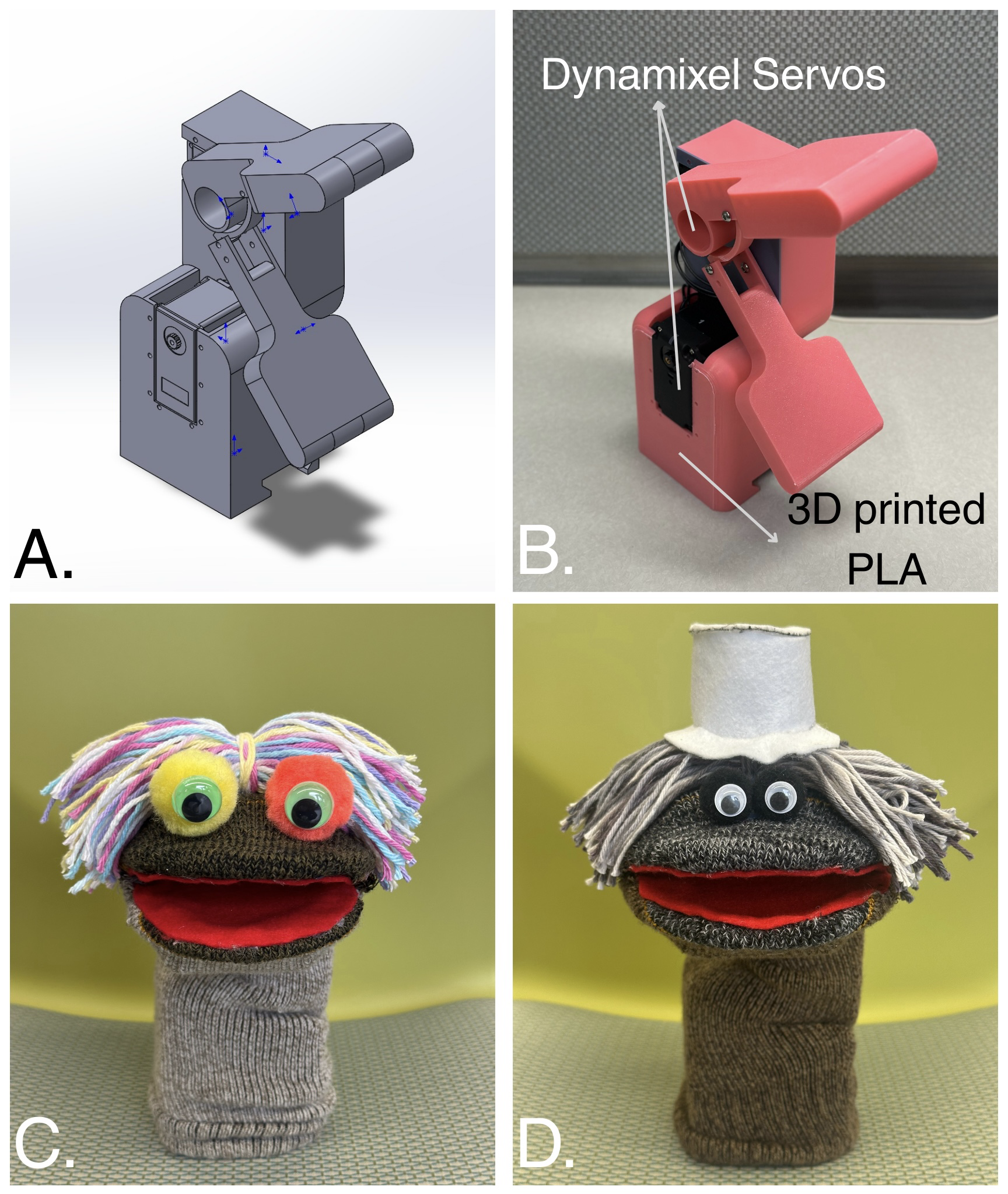}
    \Description{Four images arranged in a 2×2 grid. (Top left, A) A gray CAD rendering of a compact robotic end-effector with jointed components. (Top right, B) A pink 3D-printed robotic end-effector with visible Dynamixel servos and labeled PLA structure. (Bottom left, C) A sock puppet named Barneby with multicolored yarn hair, mismatched button eyes, and an open red mouth. (Bottom right, D) A sock puppet named Fitzwilliam with gray yarn hair, round button eyes, a small white hat, and an open red mouth.}
    \vspace{-0.5cm}
    \caption{\normalfont{\textit{Hardware and character designs for RoboTales sock puppets: A. ~CAD model of the custom 2-DoF puppet end-effector, B. ~fully assembled 3D-printed end-effector with embedded servos, C.~Barneby, a playful character, and D. ~Fitzwilliam, a reserved puppet.}}}
    \label{fig: prototyping}
\end{figure}


Two Dynamixel servomotors were selected for their power-to-size ratio and controllability: an MX-64 handles the neck actuation and an MX-28 controls the mouth movement, reducing load on the neck assembly. Because all components are modular, the end-effectors can be attached to virtually any robotic arm, making RoboTales broadly deployable across various educational environments.
\begin{figure*} [h!]
    \centering
    \includegraphics[width=\linewidth, alt={A chart displaying HRIES results across 16 individual bloxplots arranged in a grid. Each plot compares two conditions: ``puppet'' (indicated in pink) and ``green'' (indicated in green). The vertical axis represents the Likert scale from one to seven, while the horizontal axis represents the lists of various social and affective metrics (warm, likeable, trustworthy, friendly, alive, natural, real, human-like, self-reliant, rational, intentional, intelligent, creepy, scary, uncanny, and weird)}, trim={0cm 0.62cm 0cm 0.25cm},clip]{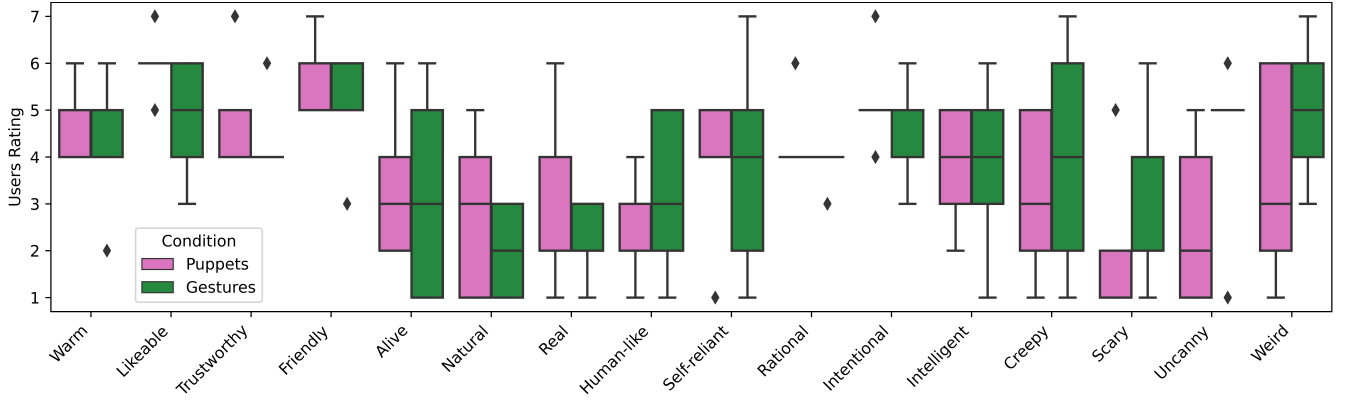}
    \Description{A chart displaying HRIES results across 16 individual bloxplots arranged in a grid. Each plot compares two conditions: ``puppet'' (indicated in pink) and ``green'' (indicated in green). The vertical axis represents the Likert scale from one to seven, while the horizontal axis represents the lists of various social and affective metrics (warm, likeable, trustworthy, friendly, alive, natural, real, human-like, self-reliant, rational, intentional, intelligent, creepy, scary, uncanny, and weird)}
    \vspace{-0.9cm}
    \caption{\normalfont{\textit{Boxplots showing the responses for all HRIES questionnaire components, separated by experimental condition. The boxes show the interquartile range (25th--75th percentiles), with the horizontal line representing the median. Whiskers extend to the most extreme values not considered outliers, while diamonds indicate outliers.}}}
    \label{fig:hries}
    \vspace{-0.4cm}
\end{figure*}

This modular design makes the end-effectors platform-agnostic. Although demonstrated on Baxter, the compact form factor, minimal wiring, and widely available components ensure that the RoboTales puppets can be adapted to most robotic platforms. By eliminating dependence on humanoid heads, complex hands, or proprietary actuation, RoboTales lowers the barrier to expressive robotic storytelling, making puppet-based building accessible to all skill levels.


Finally, the hardware additionally supports unique character expression. Two custom sock puppets were crafted to embody distinct personalities that amplify narrative engagement. The first character, \textit{Barneby}, is curious, brave, and slightly scatterbrained. The second character, \textit{Fitzwilliam}, is cautious, thoughtful, and organized. Their visual and behavioral contrasts help children recognize character roles through simple motion and design cues, reinforcing story comprehension. Figure~\ref{fig: prototyping} provides a visual representation of the puppet designs and their associated personalities.

\vspace{-0.1cm}
\subsection{Software Design}
\vspace{-0.1cm}
RoboTales software system coordinates narration, gesture generation, and embodiment control to produce expressive robotic storytelling. Because puppetry relies heavily on timing and multimodal cues, the pipeline ensures that the arm movements, mouth articulation, and visual elements unfold in synchrony with the spoken story, creating a performance that feels deliberate and character-driven.
\vspace{-0.2cm}
\subsubsection{Puppet Storytelling Mode}
In puppet mode, RoboTales uses an audio-driven actuation pipeline to animate each sock puppet's mouth in real time. The recorded narration is divided into short Root Mean Square (RMS) windows to detect speech periods, and the resulting signal is processed with a deadband and an Exponential Moving Average (EMA) filter to suppress background noise. This refined amplitude profile is then mapped to pre-calibrated minimum and maximum jaw angles, determined through Dynamixel testing, enabling expressive mouth motion without requiring speech recognition. This lightweight yet effective approach convincingly reinforces the illusion that each puppet is vocalizing its lines.

To support expressive embodiment, the robot's arm trajectories were generated using probabilistic motion primitives (ProMPs) \cite{Paraschos2013, Fabisch2024} to create varied gestures aligned with narrative beats. Rather than replaying rigid trajectories, ProMPs introduce controlled motion variation and smoothness, enabling the puppets to express their personalities through movement (e.g., Barneby's energetic gestures or Fitzwilliam's more measured, thoughtful actions). Because these gestures are time-agnostic, they operate independently of narration timing, allowing motion and dialogue to blend naturally.  

The software also supports optional visual context cues (e.g., background images and props) that educators or operators can trigger, enabling easy adaptation to classroom, library, or public demonstration environments without extra robotics infrastructure. 
\vspace{-0.5cm}
\subsubsection{Non-Puppet Storytelling Mode}
In the non-puppet mode, RoboTales shifts expressive responsibility from the puppets to Baxter itself. Instead of physical puppet articulation, the system uses a viseme-based graphical tool \cite{RhubarbLipSync} to synchronize Baxter's on-screen mouth shapes with the narration. This creates a clear impression of speech without requiring lip-syncing or conversational AI.

To maintain engagement, Baxter performs ambient conversational gestures using ProMPs. Unlike the puppet mode's more personality-driven motions, these gestures emulate the subtle, smooth, and rhythmic arm movements a human might use while speaking. To prevent over-animation, each potential gesture opportunity is evaluated and triggered with a 60\% probability, ensuring that Baxter gestures frequently enough to appear expressive, but not so often that the robot becomes distracting.


%% file: sections/user_study.tex
\section{Pilot Study Validation}
\vspace{-0.1cm}
To evaluate the efficacy of RoboTales, we conducted a small-scale between-subjects pilot study with ten participants ($M_{age} = 23.20, SD = 2.90$), who were not compensated. 
This pilot study evaluated whether RoboTales met the success criteria (Sec.~\ref{sec:goals}), motivated by gaps in prior robotic storytelling work and practical deployment constraints. Participants were randomly assigned to one of two conditions and observed Baxter perform the same five-minute story either with \textit{puppets} or using \textit{gestures only}. Afterward, participants completed the Human–Robot Interaction Evaluation Scale (HRIES; 7-point Likert) and a story-recall questionnaire. Story recall was scored using a weighted rubric across six questions of increasing difficulty (easy: 1-point; medium: 2-points; hard: 3-points). These measures evaluated emotional engagement and narrative comprehension (S1). We also assessed autonomous delivery and hardware/software reliability during live operation (S2). 

After completing the pilot study, we mounted the Barneby sock puppet onto an ABB IRB 120 arm (see Fig.~\ref{fig:baxterwithpuppets}~B.) to assess the generalizability of our hardware and software system. This transition required only a minor modification to the end-effector adapter, demonstrating the ease of mechanical integration. The full hardware workflow, from CAD design to 3D printing and installation, took approximately 45 minutes. Because audio and timing cues are preprocessed, the narrative remains plug-and-play across platforms. For arm motion, the ProMP parameters were configured for six DoF. We added a CSV parser to interface with ABB's RAPID programming environment. Together, these changes enable RoboTales to operate seamlessly on a second robot, validating the system's adaptability (S3).\looseness-1  

\begin{figure}[h!]
    \centering
    \includegraphics[width=0.55\linewidth, alt={A boxplot comparing story recall scores between two conditions: ``puppet'' (indicated in pink) and ``gestures only'' (indicated in green). The vertical axis represents the total recall score, ranging from 0 to 12, while the horizontal axis represents each condition. The puppet memory recall was higher than the gestures only condition and had less variance.}, trim={0cm 0.25cm 0cm 0.25cm}, clip]{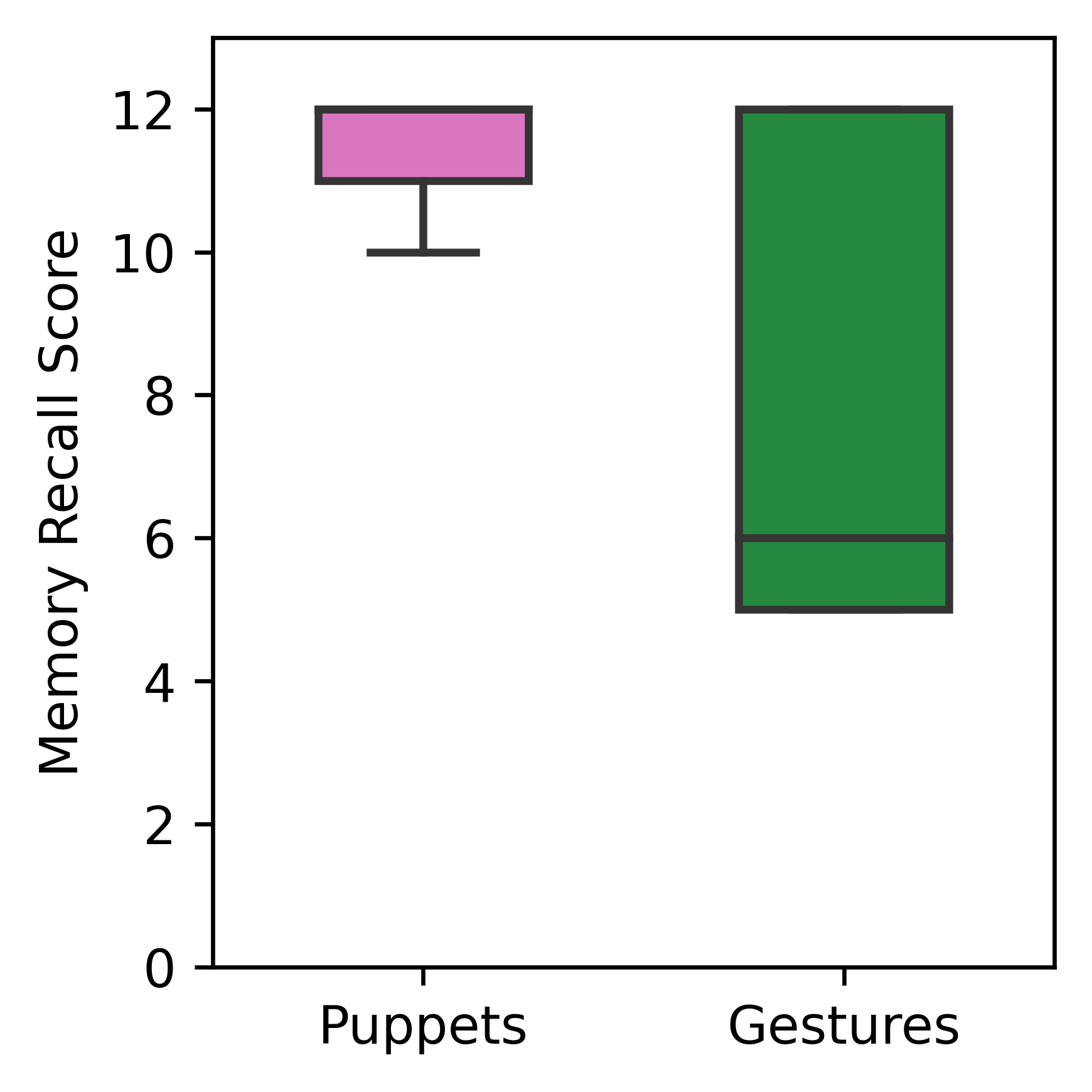}
    \Description{A boxplot comparing story recall scores between two conditions: ``puppet'' (indicated in pink) and ``gestures only'' (indicated in green). The vertical axis represents the total recall score, ranging from 0 to 12, while the horizontal axis represents each condition. The puppet memory recall was higher than the gestures only condition and had less variance.}
    \vspace{-0.5cm}
    \caption{\normalfont{\textit{Boxplots showing the distribution of memory recall scores, separated by experimental condition.}}}
    \vspace{-0.6cm}
    \label{fig:mem}
\end{figure}



%% file: sections/06_results.tex
\section{Findings and Reflections}
\vspace{-0.1cm}

The pilot study and follow-up test provide preliminary support that RoboTales meets its design success criteria. 
\vspace{-0.1cm}
\subsubsection*{\textbf{S1: Narrative Comprehension and Engagement}}
Participants in the \textit{puppet condition} assigned more favorable HRIES affective and social perception ratings and exhibited higher story recall scores than those in the \textit{gesture-only} condition. Although the pilot was not designed for statistical inference, these consistent trends indicate that the embodied puppetry cues enhanced both emotional engagement and narrative comprehension, satisfying S1. Figure~\ref{fig:hries} highlights differences in HRIES ratings, and Fig.~\ref{fig:mem} shows higher recall performance in the puppet condition.
\vspace{-0.1cm}
\subsubsection*{\textbf{S2: Autonomous and Accessible Operation}}
Across all sessions, RoboTales delivered complete stories without operator intervention. The system successfully synchronized narration, gesture primitives, and puppet articulation, confirming fully autonomous operation using off-the-shelf hardware, thereby fulfilling S2.
\vspace{-0.1cm}
\subsubsection*{\textbf{S3: Modularity and Deployability}}
The puppet end-effectors and software pipeline performed reliably throughout testing. The hardware design supports scaling by modifying the CAD and substituting smaller, low-cost servos, enabling adaptation to affordable, open-source manipulators, such as the So-100 arm. Additionally, they were successfully deployed on an ABB IRB 120 robot arm with minimal adapter redesign, confirming the platform-agnostic nature of RoboTales' embodiment system, thus satisfying S3.
\vspace{-0.1cm}
\subsubsection*{\textbf{Reflections and Next Steps}}
Future work will develop RoboTales into a unified, reusable system with open-source, extensible hardware and software that can be adapted across robots, puppet embodiments, and storytelling domains. We will evaluate RoboTales with children in its intended deployment context (classrooms, libraries, and community settings), including securing IRB approval and developing age-appropriate study protocols to assess interpretation, engagement, and comprehension. Beyond the current puppetry versus gesture-only comparisons, we will explore a broader range of movement qualities inspired by traditional puppetry by expanding ProMP libraries and systematically varying timing, amplitude, and motion style. We will also improve mouth-timing alignment and expand puppet expressivity. These efforts will advance RoboTales to a robust, child-centered storytelling system.

%% file: sections/05_empower_society.tex
\section{Empowering Society}
\vspace{-0.1cm}

RoboTales provides a \textit{socially interactive, embodied alternative} to passive media, supporting storytime in schools, libraries, and community centers where staff time is limited. By autonomously delivering puppet-driven storytelling, RoboTales helps \textit{reduce reliance on screens} while lowering the expertise needed to run literacy-oriented programming. Rather than replacing caregivers, the robot \textit{extends their reach}: a teacher, librarian, or after-school coordinator can initiate a session while attending to other duties. In doing so, RoboTales helps \textit{scale human intention} by enabling \textit{more stories for more children, with less burden on the adults who care for them.}